\documentclass{article}

\usepackage{arxiv}
\usepackage{amsmath}
\usepackage[utf8]{inputenc} % allow utf-8 input
\usepackage[T1]{fontenc}    % use 8-bit T1 fonts
\usepackage{hyperref}       % hyperlinks
\usepackage{url}            % simple URL typesetting
\usepackage{booktabs}       % professional-quality tables
\usepackage{amsfonts}       % blackboard math symbols
\usepackage{nicefrac}       % compact symbols for 1/2, etc.
\usepackage{microtype}      % microtypography
\usepackage{lipsum}		% Can be removed after putting your text content
\usepackage{graphicx}
\usepackage{natbib}
\usepackage{doi}
\usepackage{caption}
\usepackage{subcaption}

\title{Towards Agent-based Test Support Systems: An Unsupervised Environment Design Approach}

\author{ \href{https://orcid.org/0000-0002-3672-0240}{\includegraphics[scale=0.06]{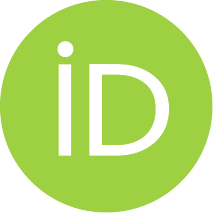}\hspace{1mm}Collins O.~Ogbodo} \\
	Dynamic Research Group\\
	School of Mechanical, Aerospace and Civil Engineering\\
        University of Sheffield\\
	Mapping Street, Sheffield, S1 3JD, United Kingdom. \\
	\texttt{coogbodo1@sheffield.ac.uk} \\
	\And
        \href{https://orcid.org/0000-0002-3433-3247}{\includegraphics[scale=0.06]{orcid.pdf}\hspace{1mm}Timothy J. ~Rogers} \\
	Dynamic Research Group\\
	School of Mechanical, Aerospace and Civil Engineering\\
        University of Sheffield\\
	Mapping Street, Sheffield, S1 3JD, United Kingdom. \\
	\texttt{Tim.rogers@sheffield.ac.uk} \\
        \And
        \href{https://orcid.org/0000-0003-4263-0513}{\includegraphics[scale=0.06]{orcid.pdf}\hspace{1mm}Mattia Dal ~Borgo} \\
	Siemens Digital Industries Software NV\\
	Interleuvenlaan 68, 3001 Leuven, Belgium.\\
	\texttt{mattia.dal\_borgo@siemens.com} \\
        \And
	\href{https://orcid.org/0000-0002-7266-2105}{\includegraphics[scale=0.06]        {orcid.pdf}\hspace{1mm}David J. ~Wagg} \\
	Dynamic Research Group\\
	School of Mechanical, Aerospace and Civil Engineering\\
        University of Sheffield\\
	Mapping Street, Sheffield, S1 3JD, United Kingdom. \\
        The Alan Turing Institute\\
        NW1 2DB, London, United Kingdom.\\
	\texttt{David.wagg@sheffield.ac.uk} \\
}

\renewcommand{\shorttitle}{\textit{arXiv} Template}

\hypersetup{
pdftitle={A template for the arxiv style},
pdfsubject={q-bio.NC, q-bio.QM},
pdfauthor={David S.~Hippocampus, Elias D.~Striatum},
pdfkeywords={Modal testing, Adaptive Design of Experiment, Reinforcement Learning, Adaptive Curriculum Learning},
}

\begin{document}
\maketitle

\begin{abstract}
Modal testing plays a critical role in structural analysis by providing essential insights into dynamic behaviour across a wide range of engineering industries. In practice, designing an effective modal test campaign involves complex experimental planning, comprising a series of interdependent decisions that significantly influence the final test outcome. Traditional approaches to test design are typically static—focusing only on global tests without accounting for evolving test campaign parameters or the impact of such changes on previously established decisions, such as sensor configurations, which have been found to significantly influence test outcomes. These rigid methodologies often compromise test accuracy and adaptability. To address these limitations, this study introduces an agent-based decision support framework for adaptive sensor placement across dynamically changing modal test environments. The framework formulates the problem using an underspecified partially observable Markov decision process, enabling the training of a generalist reinforcement learning agent through a dual-curriculum learning strategy. A detailed case study on a steel cantilever structure demonstrates the efficacy of the proposed method in optimising sensor locations across frequency segments, validating its robustness and real-world applicability in experimental settings.

\end{abstract}

% keywords can be removed

\section{Introduction}
Testing is a critical stage in every engineering design process, often consuming a significant portion of both time and budget. In structural dynamics, modal testing plays an essential role in understanding and verifying the dynamic behaviour of structures through their responses to vibrations, varying loads, or other dynamic forces. The resulting data enables engineers to optimise designs, validate control strategies, and identify potential structural weaknesses, thereby ensuring both the integrity and safety of the structure.

A crucial objective of modal testing is the validation of mathematical models that predict structural behaviour. By conducting targeted experiments, engineers can obtain critical information on parameters such as loads, damping, and other dynamic quantities. Depending on the context, these tests might be executed on reduced-scale physical models, typically for building structures, or directly on full-scale structures, as is often the case with automotive testing. For aerospace vehicles, the stakes are even higher: thorough dynamic ground tests, including ground vibration tests, are performed on aircraft and spacecraft prior to any flight \cite{craig2006fundamentals}. These tests not only confirm theoretical models but also facilitate a deeper understanding of inherent dynamics for the design and improvement of controllers. Modal testing is also used to experimentally determine the durability of structures and diagnose machinery for maintenance \cite{inman1994engineering}. Given the integral role that modal testing plays across diverse industries, designing an efficient testing strategy is important to achieve the best outcomes for downstream application processes \cite{melero2022design}.

Modal testing campaigns often consist of a sequence of global and detailed local tests that capture the essential dynamics of a given structure and, therefore, require a robust test design. Each test campaign begins with a pre-test process, which comprises a series of decisions about test environment design, detailed design of the experiment, and test setup process \cite{maia1997theoretical}. Test design decisions such as output (sensor) position on the structure have been found to significantly affect test results regardless of the test objective, test environment and frequency range of interest \cite{kelmar2024optimization, pala2023determining, woodall2023effective}. Unlike other test design decisions that are quite direct, output location decisions require an optimisation process, which in most cases is computationally intensive given the dimensions of the test structure. The complexity is further compounded when test campaigns incorporate multiple distinct test environment configurations. This highlights the need for optimal output position decision-making across changing test environment parameters without necessitating separate optimisations for each test. Furthermore, this challenge also presents itself in situations where there are unexpected changes in the test environment setup requiring a quick adaptation of the original test design.

To address these challenges, this study introduces an agent-based decision support system that facilitates optimal output position decisions across changing test environment parameters, eliminating the need for multiple computationally intensive optimisations. The proposed framework is built upon the principles of the Markov property and is solved using a reinforcement learning (RL) agent that leverages an adaptive curriculum learning strategy. We demonstrate its efficacy on an optimal sensor placement decision problem across frequency spectrum, which facilitates detailed structural characterisation within different frequency segments of interest following an initial global broadband test.

 This study is organised as follows: In \hyperref[sec: RW]{Section 2}, we introduce related works such as dual curriculum design and design of experiments in modal testing. \hyperref[sec: PF]{Section 3} details the formulation of the problem of interest. Case study results are presented in \hyperref[sec: E]{Section 4}, followed by discussions on findings in \hyperref[sec: D]{Section 5}, and conclusion in \hyperref[sec: C]{Section 6}.

\section{Related Work}\label{sec: RW}
\subsection{Dual Curriculum Design}\label{subsec: DCD}
Dual curriculum design (DCD) is considered a class of unsupervised environment design (UED) that aims to improve the generalisation capability of reinforcement learning agents across a distribution of environment levels \cite{dennis2020emergent}. Within DCD, two co-evolving teachers shape the learning experience of a student agent: one teacher actively generates new, challenging environment levels, while the other curates a pool of existing levels by prioritising those estimated to be difficult for the student \cite{jiang2021replay}. Together, this dual-teacher approach continuously refines and pushes the student policy toward higher performance. Based on the output student agent, a DCD algorithm can yield a generalist (a single agent capable of generalising across the environment level distribution) as demonstrated by methods such as Protagonist Antagonist Induced Regret Environment Design (PAIRED) \cite{dennis2020emergent,jiang2021replay}, Prioritised Level Replay (PLR) \cite{jiang2021prioritized}, and Adversarially Compounding Complexity by Editing Levels (ACCEL) \cite{parker2022evolving}. Alternatively, DCD can produce a population of specialist student agents, with each student becoming an expert in different segments of the environment level distribution, as is the case for the Paired Open-Ended Trailblazer (POET) algorithm \cite{wang2019paired}. Although DCD was originally developed within the reinforcement learning community, it presents an elegant framework for adaptive design of experiments; this suitability is further discussed in \hyperref[subsec: UPOMDP]{Section 3.1}

\subsection{Design of Experiment in Modal Testing}\label{subsec: DEDT}
Designing a modal test requires first establishing a clear objective, which may range from determining resonance frequencies to updating finite element (FE) models, as described by \cite{maia1997theoretical}. This is followed by the determination of the number of exciters and sensors and their corresponding locations on the structure. In practice, when a reliable FE model is available, it facilitates these design decisions; otherwise, engineers must rely on a trial-and-error approach, leveraging experience and judgement in the absence of prior knowledge of the structure's dynamic behaviour \cite{maia1997theoretical}. The mathematical foundation for the sensor placement problem has been well established in a series of journal papers, each contributing either an objective function—such as effective independence by \cite{kammer1991sensor}, Fisher information matrix metrics by \cite{middleton1960introduction}, effective independence driving point residue by \cite{papadopoulos1998sensor}, and kinetic energy measures by \cite{heo1997optimal}—or an optimisation methodology, either deterministic, stochastic or data driven \cite{paris2021robust}, \cite{hassani2023systematic}. In practice, these sensor placement problems are typically solved as static problems, often focusing on a global broadband test after which the test design (particularly the sensor locations) remains unchanged throughout the entire test campaign. 

This study, therefore, makes a major contribution: we introduce a novel framework that combines physics-based simulation with data-driven reinforcement learning techniques, enabling an adaptive test design strategy that overcomes the limitations of traditional static approaches. By formulating the experiment design problem as a Markov decision process (MDP), our framework inherently captures the changing parameters of modal testing environments. 

\section{Problem Formulation}\label{sec: PF}
Building on the formulation established in \cite{ogbodo2025adaptive}, the output location decision-making problem can be reformulated as a sequential decision-making process. This reformulation renders the classical MDP methodology particularly suited for establishing a robust mathematical framework for a static optimal sensor placement problem. However, in this study, the challenge is exacerbated by the need to tailor these decisions across a range of test designs, each distinguished by a unique test design parameter. To accommodate this additional complexity, we leverage the Underspecified Partially Observable Markov Decision Process (UPOMDP), which is an extension of the MDP framework. Compared to \cite{ogbodo2025adaptive}, in this formulation, the state space, transition dynamics, and reward function are not only governed by the intrinsic properties of the test environment but are also modulated by the test design parameters. Consequently, the UPOMDP naturally captures the variability in experimental configurations, enabling sensor placement decisions to adapt dynamically to changes in the testing environment.

\subsection{Underspecified Partially observable Markov Decision Process} \label{subsec: UPOMDP}
As previously established, because the problem is inherently sequential, we model it using the MDP framework which is defined by the tuple $\langle \mathcal{S}, \mathcal{A}, \mathcal{T}, \mathcal{R}, \gamma \rangle$ where 
\begin{itemize}
    \item $s \in \mathcal{S}$ is the set of states.
    \item $a \in \mathcal{A}$ is the set of actions.
    \item $\mathcal{T}: p(s' | s,a) := Pr\{S_{t+1} =s' | S_{t} = s, A_{t} = a\} $ is the state transition probability matrix, representing the dynamics of the test environment.
    \item $\mathcal{R}: R(s, a)$ is the reward function.
    \item $\gamma \in [0,1]$ is the discount factor.
\end{itemize}
%%%The problem also naturally satisfies the Markov assumption, denoted as
%%%\begin{equation}
%%%    Pr\{S_{t+1} = s' | S_t = s, A_t =a \} = Pr\{S_{t+1} = s' | S_t = s_{t, t-1,...,0}, A_t =a \}
%%%    \label{equ:Markov Assumption}
%%%\end{equation}

Considering the adaptive structure of the problem, we utilise UPOMDP proposed by \cite{dennis2020emergent} for unsupervised environment design in reinforcement learning. UPOMDP improves the MDP framework in two limiting areas; namely, the first is the assumption of full observability of the agent, which is addressed by the Partially Observable Markov Decision Process (POMDP) by introducing an observation function that maps the unknown true state to noisy sensor observations, and the second is that the transition and reward function are fixed across all state-action pairs throughout training, which in practice is not true given that the agent may experience a variation of the environment not seen during training, resulting in the need for robustness across changes in environments \cite{parker2022evolving}. The UPOMDP addresses the second issue by introducing an environment parameter unique to each environment and is defined as a tuple $\langle \mathcal{S}, \mathcal{O}, \mathcal{A}, \mathcal{T}, \mathcal{R}, \mathcal{I}, \Theta, \gamma \rangle$ where, in addition to the previous definition 
\begin{itemize}
    \item $o \in \mathcal{O}$ is the set of observations
    \item $\theta \in \Theta$ environment parameter of interest
    \item $\mathcal{I}: \mathcal{S} \rightarrow \mathcal{O} $ is the observation function
\end{itemize}

The parameter $\Theta$ may vary between episodes and is embedded within the transition function $\mathcal{T} : \mathcal{S} \times \mathcal{A} \times \Theta \rightarrow \mathcal{S}$. Consequently, each trajectory is governed by the specific parameter associated with the rollout environment. Analogous to \cite{dennis2020emergent}, the objective is to learn a policy that generalises across the distribution of these environment parameters.

Originally formulated for modelling underspecified RL environments for UED problem, the UPOMDP formulation offers a rich mathematical foundation for the described problem. In adaptive experimental design, the testing environments are parameterised by the test design variable, much like an RL setting, where each environment variant is parameterised by environment features  (such as the agent's start and target locations and block positions within the grid for Minigrid environments). These parameters collectively shape the environment's complexity, which in turn determines its level. Here, the notion of "environment feature" in UPOMDP, which modulates the difficulty or characteristics of the generated environments, corresponds directly to an experimental design parameter that defines key aspects of the testing setup. $\Theta$, therefore, parameterises the testing environment controlling features such as simulated damage conditions, frequency segments, structure geometry or boundary conditions (type or stiffness).

\subsection{Adaptive Test Environment}\label{subsec: ATE}
In a previous work, we introduced an RL environment framework for adaptive sensing in digital twins \cite{ogbodo2025adaptive}. The current study builds on this environment framework, where the state space represents output positions within a discretised grid, encoded as a multi-binary vector indicating sensors mounted at specific locations. The action space is multi-discrete, allowing for the selection of a sensor and steering it in specific directions. The transition matrix is deterministic, ensuring that sensor movements always lead to predictable configuration updates, without randomness in state transitions. The discount factor is chosen to balance short-term rewards with long-term optimisation, guiding the agent toward globally optimal sensor configuration rather than immediate modification. 

The reward function as presented in \cite{ogbodo2025adaptive} measures the improvement in sensor configuration resulting from a change in the position of one sensor at each step. This improvement is an information entropy measure, which quantitatively measures the probabilistic uncertainty in the estimated model parameter given a data stream from the sensor location. Particularly, the Fisher information matrix (FIM) forms the basis for the reward function and is given as
\begin{equation}
\mathbf{Q}(\mathbf{L}, \boldsymbol{\Sigma}) =(\mathbf{L} \boldsymbol{\Phi})^T (\mathbf{L} \boldsymbol{\Sigma} \mathbf{L}^T)^{-1} (\mathbf{L}\boldsymbol{\Phi})
    \label{eq:FIM-phi}
\end{equation}
where $\mathbf{L}$ is a binary observation matrix that specifies the sensor location being monitored, $\boldsymbol{\Phi}$ is the structure's mode shape matrix, and $\boldsymbol{\Sigma}$ is the covariance matrix of the model parameter and models the spatial correlation between sensors defined as
\begin{equation}
    \Sigma_{ij} = \exp\left(-\frac{\upsilon_{ij}}{\upsilon}\right) \frac{\boldsymbol{\psi}_i^\top \boldsymbol{\psi}_j}{N_M},
\end{equation}
given that $\mathbf{\boldsymbol{\psi}}_i = \psi_{\textit{k},i} \mid \textit{k} = 1, \ldots, K $ and $\mathbf{\boldsymbol{\psi}}_j = \psi_{\textit{k},j} \mid \textit{k} = 1, \ldots, K $ which are evaluated as:
\begin{equation}
    \psi_{k,i} = \frac{|\phi_{k,i}|}{\max(|\phi_{k,i}|, |\phi_{k,j}|)}, \quad \psi_{k,j} = \frac{|\phi_{k,j}|}{\max(|\phi_{k,i}|, |\phi_{k,j}|)},
\end{equation}
while \( \upsilon \) is defined as the ratio of the greatest distance across all DOFs to the total number of sensors, and $\upsilon_{ij}$ is the spatial correlation. The terms \( \phi_{k,i} \) and \( \phi_{k,j} \) are the mode shapes at positions \( i \) and \( j \) for mode \( k \), respectively. Accordingly, the immediate reward at each step is defined as the difference between successive evaluations of the determinant of the Fisher information matrix (the reward metric), computed as follows:
\begin{equation}\label{eq: reward}
R := det(\mathbf{Q}(\mathbf{L'},\boldsymbol{\Sigma}))_{current} - det(\mathbf{Q}(\mathbf{L},\boldsymbol{\Sigma}))_{previous}
\end{equation}

For UPOMDP, the reward function as described in equation \ref{eq: reward} becomes a function of the environment parameter $\Theta$ and is given as
\begin{equation}
R(\Theta) := det(\mathbf{Q^\theta}(\mathbf{L'},\boldsymbol{\Sigma}))_{current} - det(\mathbf{Q^\theta}(\mathbf{L},\boldsymbol{\Sigma}))_{previous}
\end{equation}\label{eq: upomdp reward}

\subsection{Unsupervised Environment Design}
\label{subsec: UED}
In this section, we introduce the UED framework, a curriculum learning approach for addressing the critical challenge of generalisation beyond trained environments in RL. Unlike traditional curriculum learning methods that require expert-designed progressions, UED autonomously generates environment distributions that promote robust policy learning. As defined by \cite{dennis2020emergent}, UED involves leveraging underspecified environments to generate a distribution of well-defined environments that facilitate continuous policy learning. The primary objective of UED is to construct a curriculum consisting of a series of levels that define the complexity of training environments, effectively supporting the ongoing learning of a specific student agent's policy, ensuring adaptability and robustness across all levels. This environment parameter curation process is performed by a teaching agent (teacher) following a specific decision rule. The minimax regret decision rule has proven effective for this process due to its ability to reliably select successful student policies in tasks with a well-defined notion of success and failure \cite{dennis2020emergent}. In the UED approach, the regret is defined as the difference in expected return between the current policy and the optimal policy: 
\begin{equation}\label{eq: Regret}
    REGRET^{\theta}(\pi, \pi^*) = V^{\theta}(\pi^*) - V^{\theta}(\pi)
\end{equation}
where $\pi$, $\pi^*$ and $V$ are the current policy, optimal policy, and value function, respectively. The teacher generates levels by maximising a regret-based utility function defined as
\begin{equation}\label{eq: utility}
    U^R_t(\pi,\theta) = \operatorname*{argmax}_{\pi^* \in \Pi}  \{REGRET^{\theta}(\pi, \pi^*)\}
\end{equation}
where $\Pi$ is the set of possible student policies. At Nash equilibrium \cite{daskalakis2009complexity} in the learning process, the resulting student converges to a minimax regret policy given $\Pi$ and a set of possible sequences of environment parameters $\Theta$ \cite{parker2022evolving}. At this stage, the student policy is then given as
\begin{equation}\label{eq: student policy}
    \pi = \operatorname*{argmin}_{\pi^* \in \Pi} \{ \operatorname*{max}_{\theta,\pi^* \in \Theta,\Pi}  \{REGRET^{\theta}(\pi, \pi^*)\}.
\end{equation}

Due to the difficulty in computing equation \ref{eq: Regret}, an efficient approximate solution proposed by \cite{jiang2021prioritized} in their UED framework PLR is used \cite{parker2022evolving}. In PLR, the student agent is trained either by sampling a new, unseen environment from a distribution $P_{new}$ or by replaying a previously encountered environment from a replay distribution $P_{replay}$ with a probability $d \sim P_D$ (where $P_D$ typically follows a Bernoulli distribution). Environments are replayed from $P_{replay}$ based on their regret score, which represents their learning potential and how long ago they have been previously sampled. $P_{replay}$ is defined as: 
\begin{equation}
    P_{replay} = (1-\rho) P_s + \rho P_C
\end{equation}\label{eq: replay dist}
here $P_s$ denotes the scoring distribution, while $P_c$ represents the staleness distribution. $P_s$ quantifies the student agent's performance on previously encountered environments using the averaged Generalised Advantage Estimation (GAE) over the most recent trajectory. The corresponding scoring function is defined as:
\begin{equation}
    S_i = \text{score}(\tau, \pi) = \frac{1}{T} \sum_{t=0}^{T} \left| \sum_{k=t}^{T} (\gamma \lambda)^{k-t} \delta_k \right|
\end{equation}\label{eq: score}
where $\lambda$ is the generalised advantage estimation and $\delta_k$ is the $k$-step temporal-difference error at time step $t$. The scoring distribution is therefore evaluated as
\begin{equation}
    P_S(l_i|\Lambda_{seen},S) = \frac{h(S_i)^{1/\beta}}{\sum_{j} h(S_i)^{1/\beta}}
\end{equation}\label{eq: replay distribution}
where $h(S_i) = 1/rank(S_i)$ maps differences in environment scores to corresponding differences in prioritisation given the ranks of the environment score $S_i$ amongst all scores when arranged in descending order. $\beta$ is a temperature parameter that controls the influence of $h$ on the resulting distribution. On the other hand, the $P_C$ models a notion of "updatedness" which guarantees that stale environments are replayed, preventing them from drifting far off-policy. $P_C$ is evaluated as
\begin{equation}
    P_C(l_i|\Lambda_{seen},C,c) = \frac{c-C_i}{\sum_{C_j \in C} c-C_j}
\end{equation}
here $c$ and $C_i$ are the global total training episodes and the number of times environment $l_i$ has been sampled. Hence, $c-C_i$ models the staleness of the environment of interest $l_i$.

Although PLR results in policies with strong generalisation, it struggles to generate more complex environments necessary for robustness due to its restriction to randomly sampled levels, which becomes inefficient in high-dimensional spaces, making it increasingly difficult to find challenging environments as the agent advances in capability \cite{parker2022evolving}. This problem is addressed by ACCEL which was introduced by \cite{parker2022evolving}. ACCEL extends PLR by including an evolutionary mutation operator. After replaying a challenging level (as in PLR), ACCEL can mutate (modify an aspect of the environment) that environment to create a new one under the assumption that regret changes gradually as environment parameters $\Theta$ are modified. This assumption allows small edits to high-regret environments to produce new environments with similar regret values, helping uncover challenging environments efficiently. By leveraging this smooth variation, ACCEL enhances the curriculum by systematically increasing complexity rather than relying on random sampling, making it a robust approach for UED. The mutated environment is then added back to the replay buffer, allowing the curriculum to evolve \cite{parker2022evolving}.

\begin{figure}[ht]
    \centering
    \includegraphics[width=1\linewidth]{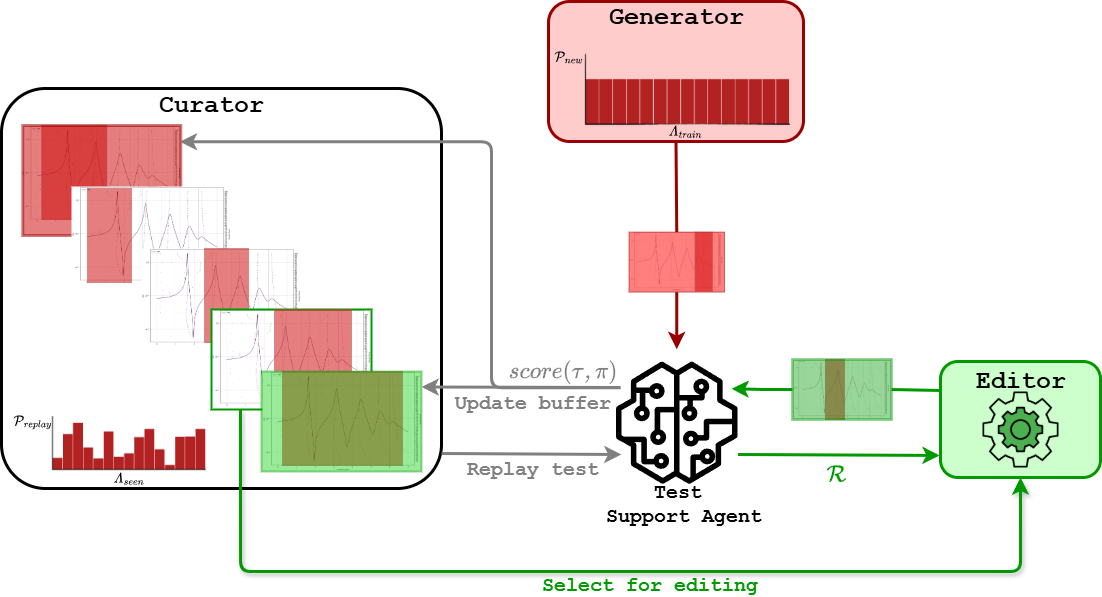}
    \caption{Adapted ACCEL framework for frequency spectrum-based environment parameter. The generator uniformly samples test environments for evaluation. Environments that exhibit high regret are then transferred to a replay buffer, from which the curator selects them for further training of the student. After the training phase, the regret values for these replayed environments are adjusted and re-evaluated, refining their selection for subsequent training iterations ( figure inspired by \cite{parker2022evolving}).}
    \label{fig: ACCEL}
\end{figure}

\begin{figure}[ht]
    \centering
    \includegraphics[width=1\linewidth]{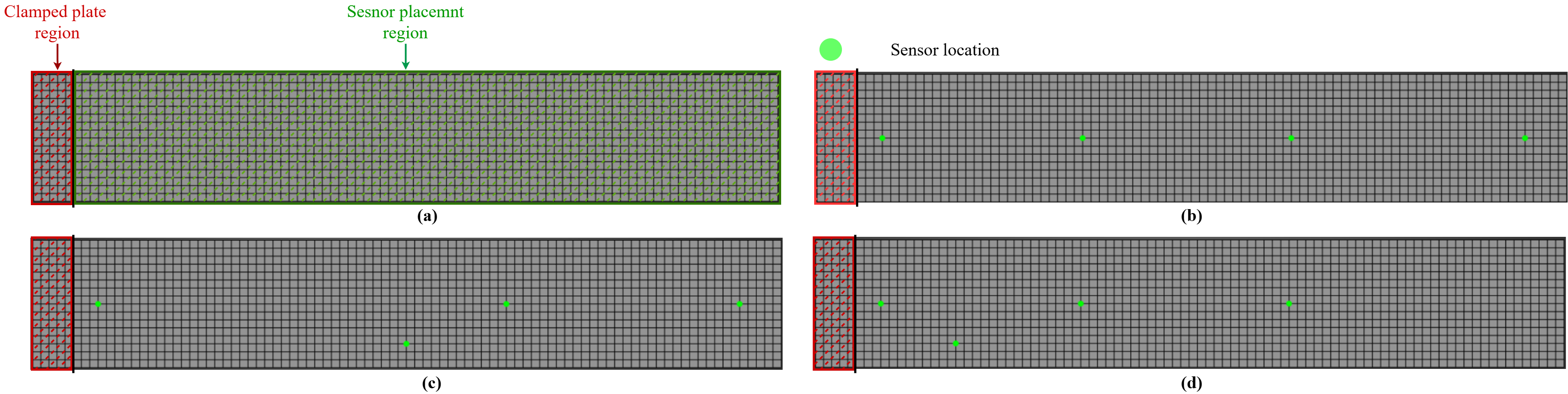}
    \caption{ACCEL mutation for a clamped cantilever with four sensors showing (a) Clamped and sensor placement region, (b) Sensor default initialisation location, (c) and (d) Randomly selected sensor moved to a random location within the sensor placement region.}
    \label{fig: ACCEL Mutation}
\end{figure}

By adopting the ACCEL framework as shown in Fig. \ref{fig: ACCEL}, a generalist student agent can be trained to make optimal output location decisions across the distribution of the test environment parameter. The mutation of designed test environments involves a change in the start location of a randomly sampled sensor to a random position within the prespecified sensor placement region during environment initialisation or reset, as shown in Fig. \ref{fig: ACCEL Mutation}. 

\section{Experiment}\label{sec: E}
We now demonstrate the effectiveness of the proposed framework on a case study. The student agent is a Long Short-Term Memory (LSTM) \cite{hochreiter1997long} trained using proximal policy optimisation (PPO) \cite{schulman2017proximal}. 

\subsection{Case Study: Clamped Cantilever Structure}\label{subsec: CC}
\begin{figure}[ht]
    \centering
    \includegraphics[width=1.0\linewidth]{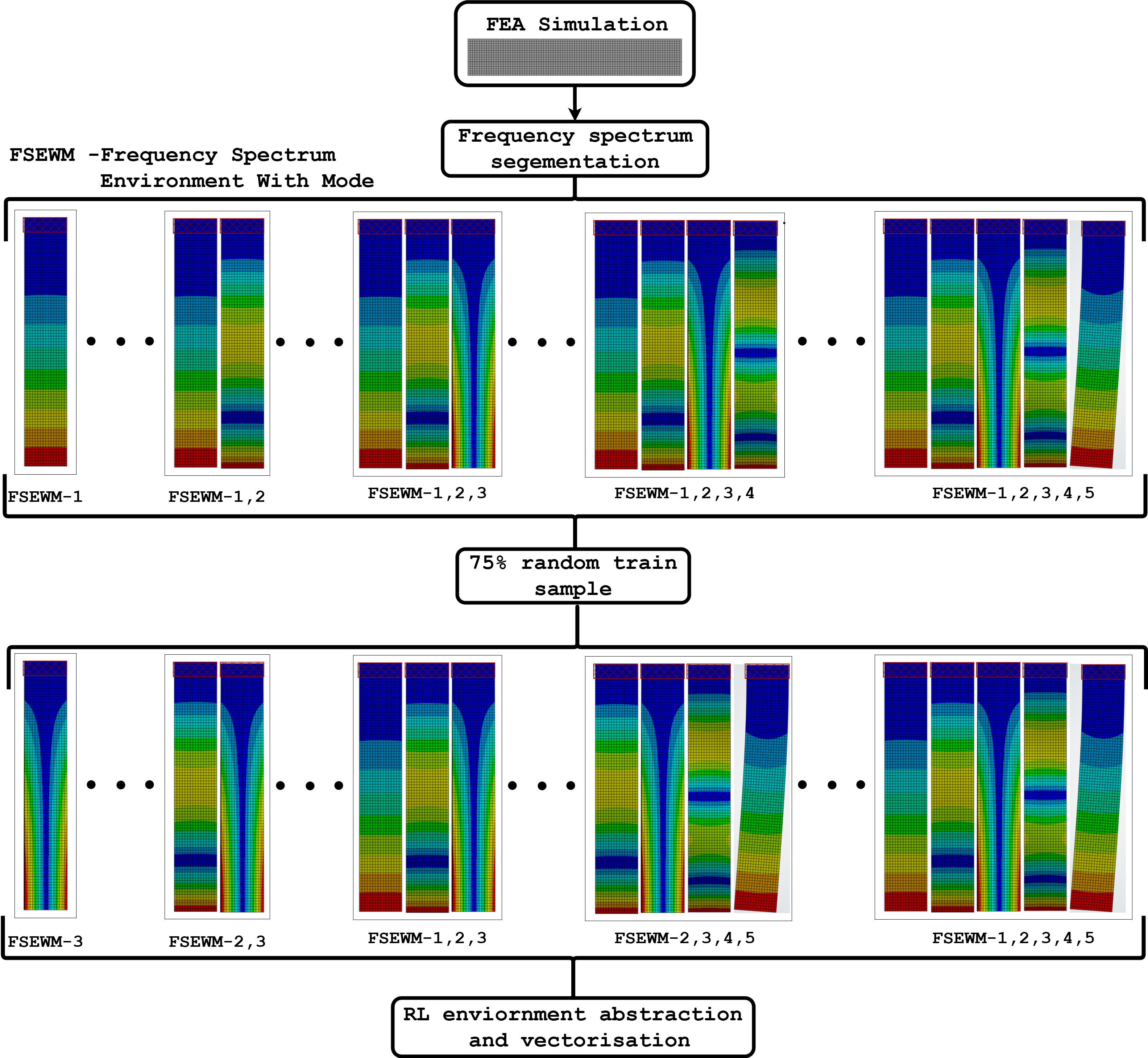}
    \caption{Training environment generation process. Training environment generation begins with a single FEA modal simulation. The resulting global frequency spectrum is then partitioned into all possible contiguous mode subsequences. For the five modes considered, the environment set includes $l_i \in \mathbf{\Lambda} \mid \{(1), (2), \ldots, (1,2), (2,3), \ldots, (1,2,3), (2,3,4), \ldots, (1,2,3,4,5)\}$. From this set, 75\% of the total environments are randomly selected to form the training set. Finally, we create the RL environment from them and vectorise for multiprocessing.}
    \label{fig: tain env}
\end{figure}

In this case study, the goal is to train a generalist agent capable of making optimal sensor location decisions for different segments of the frequency spectrum of interest. The physical structure consists of a steel cantilever plate with dimensions of 447 mm in length, 76.2 mm in width, and 3 mm in thickness. The plate is clamped at one end at a depth of 24 mm from the end. The first five modes are considered, and a total of five sensors are utilised. It is important to highlight that the agent focuses on modes that exist within the frequency range of interests and therefore positions the sensors to maximise the information-theoretic reward function in \hyperref[subsec: DEDT]{Section 3.2}. The training environment generation process, illustrated in Fig. \ref{fig: tain env}, involves a single FEA modal analysis simulation from which 15 environments are derived. These environments represent all continuous frequency segments of the global frequency spectrum by capturing every contiguous subsequence of the five modes. 11 environments are then randomly sampled as the training set from which mutations are generated. 

We first demonstrate that the student agent is capable of learning a decision-making policy necessary for optimal sensor placement within the trained environments. The agent undergoes training for 20 million environment steps, corresponding to 4,882 policy updates. The specific hyperparameter configurations used in this training process are detailed in Appendix \ref{subsec: H}. Before commencing full training, we performed an ablation study—presented in Appendix \ref{sec: AS}—to isolate and investigate the influence of the mutation settings used within the ACCEL framework. This preliminary investigation ensures a better understanding of the agent's performance sensitivity to mutation strategy and hyperparameters.

\begin{figure}[ht]
    \centering
    \includegraphics[width=1.0\linewidth]{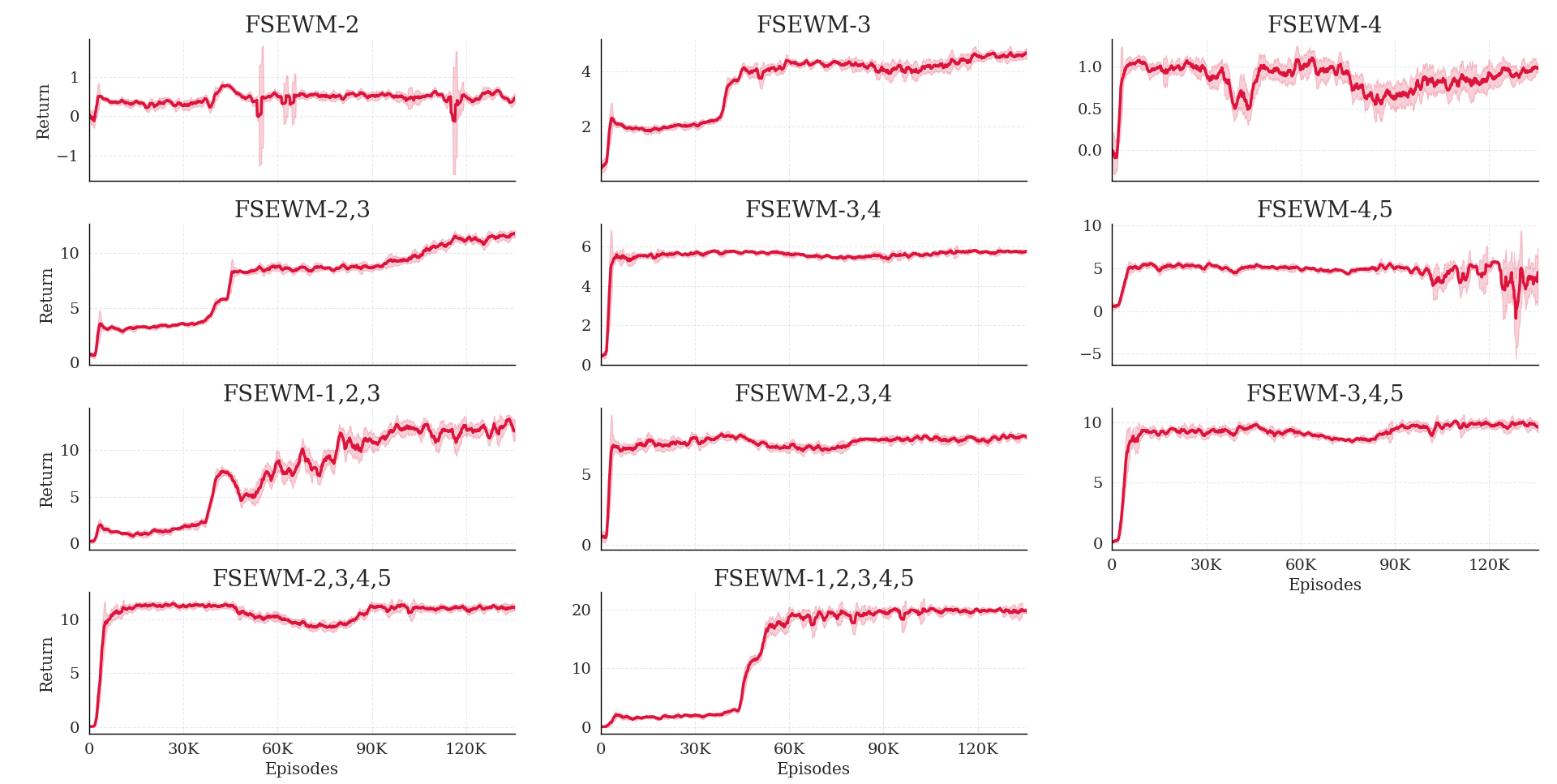}
    \caption{Student agent training performance trajectory on seen environments. Curve is smoothed with a moving average of 10 points and standard deviation (shaded region).}
    \label{fig: agent performance on trained environment}
\end{figure}

To evaluate the agent’s learning progression, we conducted an in-training test across the designed suite of environments that captures distinct segments of the global frequency spectrum. The performance trajectory, visualised in Fig. \ref{fig: agent performance on trained environment}, reveals a consistent increase in cumulative reward and policy stability, thereby confirming the agent's growing ability to generalise across these environments. This demonstrates that the student agent successfully internalises the objective of selecting sensor configurations that maximise the expected information-theoretic reward, even under dynamic test conditions.

\begin{figure}[ht]
    \centering
    \includegraphics[width=0.7\linewidth]{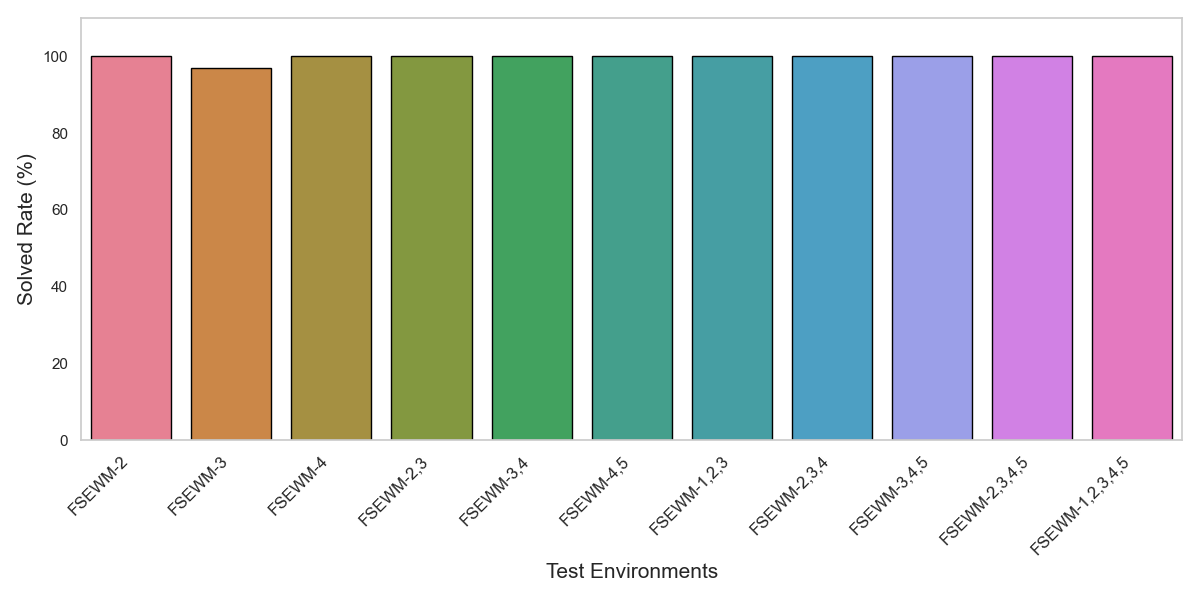}
    \caption{Student agent evaluation performance on the trained test environment for 100 episodes. The solved rate indicates the proportion of the evaluation episodes for which the agent outperforms the effective independence baseline.}
    \label{fig: solve rate}
\end{figure}

We further evaluate the performance of the trained student agent by benchmarking it against the widely used effective independence sensor placement method \cite{kim2024effective}, which serves as a baseline. The trained agent is evaluated for 100 independent episodes within the train environment, and the results are presented in Fig. \ref{fig: solve rate}. To quantify effectiveness, we define solved rate as the proportion of episodes in which the agent achieves a score higher than the baseline. For example, a solved rate of 0.7 implies that the agent outperforms the baseline in 70\% of the evaluation episodes. In addition to the solved rate, we report the mean and standard deviation of the agent’s reward scores across the evaluation set to characterise its stability and consistency. Table \ref{tab: student-agent-training} presents a detailed comparison of the agent’s reward metrics relative to those of the effective independence baseline, clearly illustrating the agent’s advantage across all evaluation environments. 

\begin{table}[ht]
\centering
\caption{Mean reward metric performance compared to the effective independence baseline for 100 evaluation episodes in the trained environment with standard deviation.}
\label{tab: student-agent-training}
\begin{tabular}{lcc}
    \toprule
    Test environment      & Student agent & Effective independence \\
    \midrule
    FSEWM-2               & $\mathbf{1.75 \pm 0.13}$  & 1.15 \\
    FSEWM-3               & $\mathbf{4.68 \pm 0.38}$  & 3.81 \\
    FSEWM-4               & $\mathbf{2.43 \pm 0.27}$  & 1.19 \\
    FSEWM-2,3             & $\mathbf{12.67 \pm 1.17}$ & 3.55 \\
    FSEWM-3,4             & $\mathbf{5.60 \pm 0.22}$  & 3.47 \\
    FSEWM-4,5             & $\mathbf{5.98 \pm 1.05}$  & 1.35 \\
    FSEWM-1,2,3           & $\mathbf{15.03 \pm 1.36}$ & 4.67 \\
    FSEWM-2,3,4           & $\mathbf{7.97 \pm 0.23}$  & 4.72 \\
    FSEWM-3,4,5           & $\mathbf{10.22 \pm 0.39}$ & 3.91 \\
    FSEWM-2,3,4,5         & $\mathbf{11.14 \pm 0.78}$ & 4.32 \\
    FSEWM-1,2,3,4,5       & $\mathbf{19.73 \pm 1.14}$ & 6.96 \\
    \bottomrule
\end{tabular}
\end{table}

\begin{figure}
    \centering
    \includegraphics[width=1.0\linewidth]{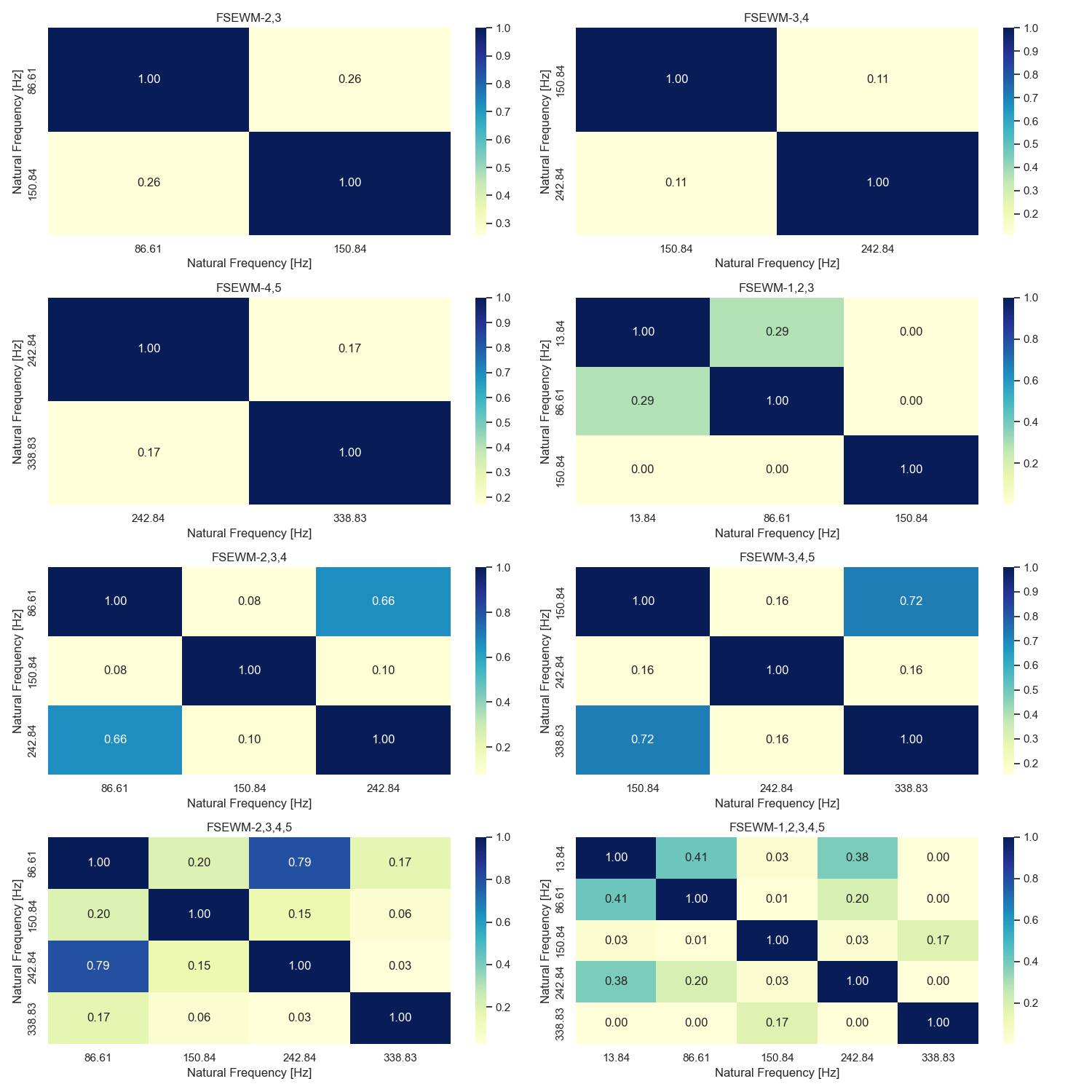}
    \caption{MAC plot for predicted sensor configuration in trained environment with two or more modes showing quality of mode independence.}
    \label{fig: global trained env MAC}
\end{figure}

In addition to reward-based results, we also assess the modal distinctiveness of the resulting sensor configurations for trained environments containing two or more modes using the Modal Assurance Criterion (MAC). The results are presented in Fig. \ref{fig: global trained env MAC}. The MAC plot serves as a diagnostic tool to evaluate the linear independence of the mode shapes identified by the selected sensor locations. According to \cite{maia1997theoretical}, an optimal sensor configuration should exhibit high diagonal dominance in the MAC matrix, with minimal off-diagonal values—an indication of low spatial correlation and minimal modal aliasing. Given that only five sensors were considered, Fig. \ref{fig: global trained env MAC} shows minimal off-diagonal values.

The MAC results further substantiate that the agent not only learns to outperform traditional placement heuristics in terms of reward optimisation but also preserves the physical interpretability and distinctiveness of the structural modes. This supports the potential of the proposed agent-based approach as a robust alternative for experimental modal design under changing test environment parameters.

\subsection{Global Performance}
\begin{figure}[ht]
    \centering
    \includegraphics[width=0.7\linewidth]{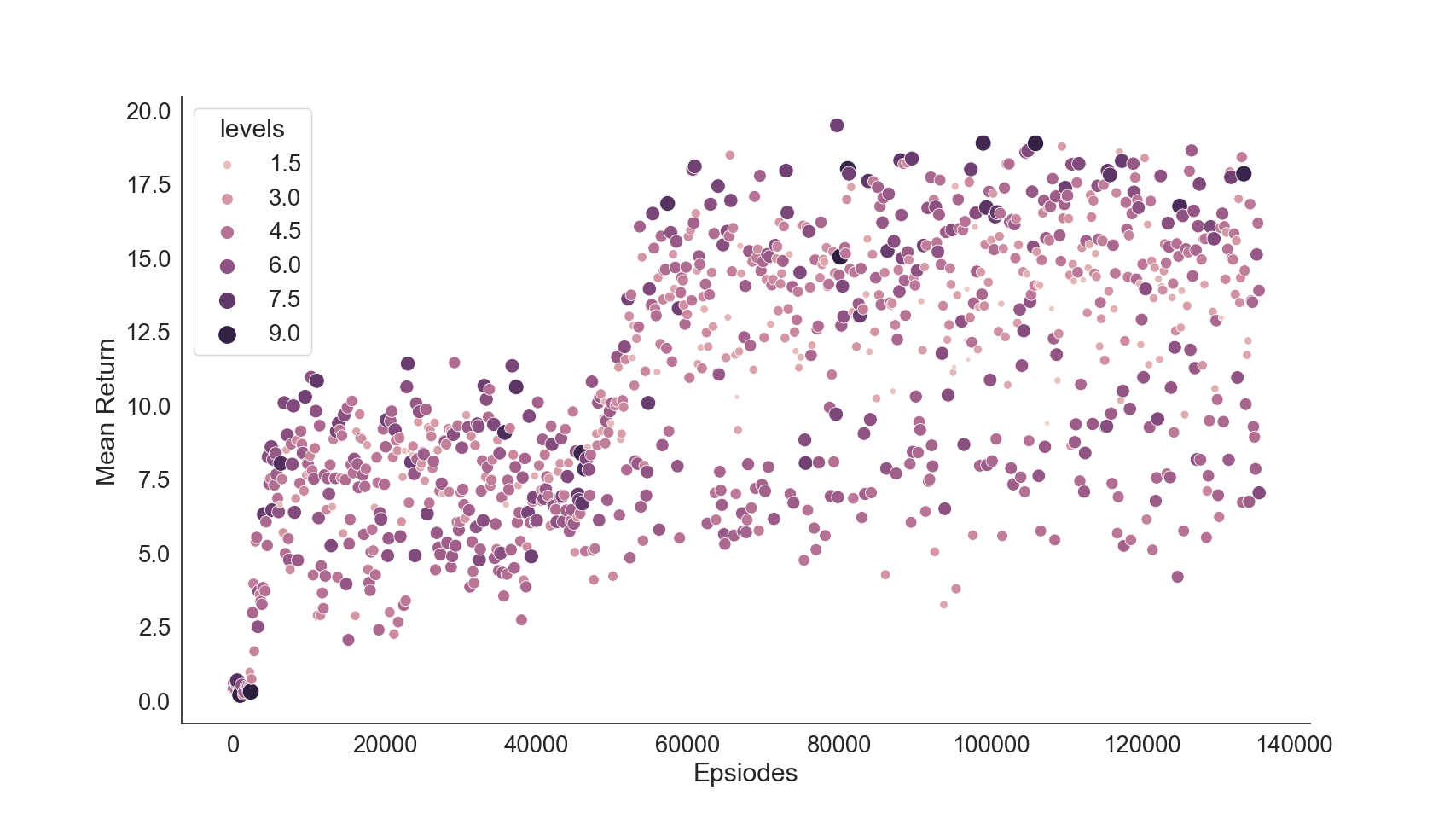}
    \caption{Student agent mean performance across environment levels. Visualised levels are averaged across parallel vectorised training environments.}
    \label{fig: agent mean level performance}
\end{figure}

We further analyse the global performance of the agent during training across all levels. As previously mentioned, levels define environment complexity. In our case, after the selection of the training set (cf. Fig. \ref{fig: tain env}), each environment is assigned a level index, beginning at zero and incrementing sequentially in the ordered set of randomly sampled training environments, i.e.,
\begin{equation}
\begin{split}
    \mathbf{\Lambda} \ni \Lambda_{train} \mid &\quad \{(2): 0, (3): 1, (4): 2, (2,3): 3, (3,4): 4, (4,5): 5, \\
    &\quad (1,2,3): 6, (2,3,4): 7, (3,4,5): 8,(2,3,4,5): 9, (1,2,3,4,5): 10\}
\end{split}
\end{equation}
where assigned levels indicate increasing environment complexity, driven both by higher-order modes with more intricate mode shapes and by the cumulative inclusion of additional modes in each environment (in this order of priority). Fig. \ref{fig: agent mean level performance} illustrates the evolution of agent performance during training across different levels, revealing a monotonic increase in mean reward with environment complexity. 

\subsection{Zero-shot Transfer}
Beyond the evaluation within trained test environments, we assess the agent’s capability to generalise to unseen test environments that represent distinct segments of the global frequency spectrum. This evaluation reflects the agent’s ability to perform zero-shot transfer across out-of-distribution environments, a highly desirable property, especially in practical scenarios where the frequency spectrum is large and encompasses numerous modes. As the number of possible test environments generated through spectral segmentation increases, exhaustive training across all configurations becomes computationally prohibitive. Consequently, the agent’s ability to train on a representative subset and maintain high performance across the full test suite is a critical requirement for scalable deployment.

\begin{figure}[ht]
    \centering
    \includegraphics[width=0.8\linewidth]{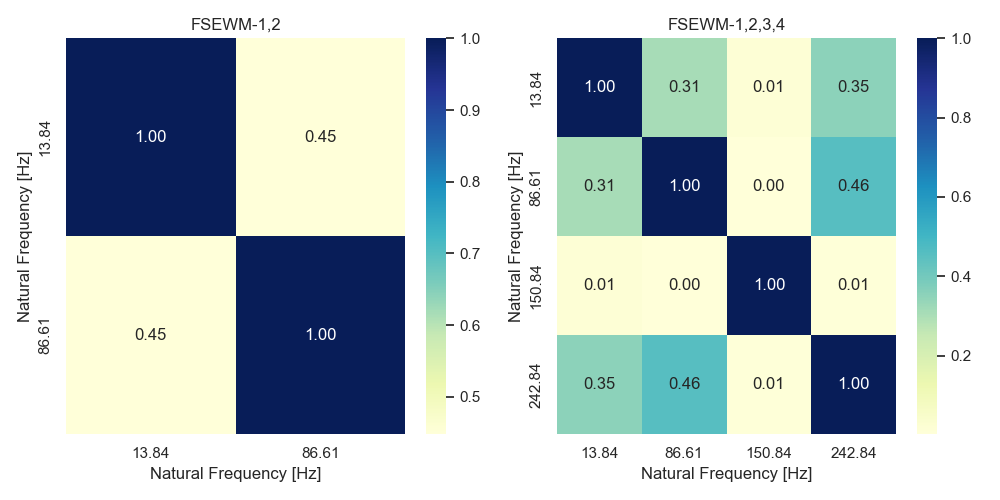}
    \caption{Agent predicted sensor configuration MAC plot for $FSEWM-1,2$ and $FSEWM-1,2,3,4$ showing quality of mode independence.}
    \label{fig: zero-shot MAC}
\end{figure}

To this end, we benchmark the agent’s performance in these unseen test environments against the baseline. As shown in Table \ref{tab: student-agent-out-of-distribution}, the student agent continues to outperform the baseline across most cases, reinforcing its capacity for generalisation for this problem. It is worth noting that the agent exhibits a lower performance score in the $FSEWM-5$ environment, which is consistent with expectations. This is attributed to the use of unidirectional sensors in the current study, whereas Mode 5, as depicted in Fig. \ref{fig: tain env}, corresponds to an out-of-plane vibration mode that is inherently more challenging to capture using such sensors. Importantly, the agent consistently identifies sensor configurations that maintain spatial decorrelation and modal distinctiveness, as evidenced in the MAC plots for $FSEWM-1,2$ and $FSEWM-1,2,3,4$ presented in Fig. \ref{fig: zero-shot MAC}. These results confirm the robustness of the learned policy and highlight the efficacy of the adaptive training framework in handling variability across diverse test scenarios.

\begin{table}[ht]
\centering
\caption{Mean reward metric performance compared to the effective independence baseline for 100 evaluation episodes in out-of-distribution test environments with standard deviation.}
\label{tab: student-agent-out-of-distribution}
\begin{tabular}{lcc}
    \toprule
    Test environment      & Student agent & Effective independence \\
    \midrule
    FSEWM-1               & $\mathbf{1.33 \pm 0.08}$  & 1.13 \\
    FSEWM-5               & $2.12 \pm 0.36$  & $\mathbf{3.73}$ \\
    FSEWM-1,2               & $\mathbf{2.30 \pm 0.24}$  & 1.24 \\
    FSEWM-1,2,3,4             & $\mathbf{20.37 \pm 2.51}$ & 16.36 \\
    \bottomrule
\end{tabular}
\end{table}

\section{Discussion}\label{sec: D}
The RL approach (particularly PPO) offers a significant advantage in addressing optimisation problems in high-dimensional spaces. This capability is especially pertinent when dealing with large-scale structures such as aircraft, buildings, and bridges, or with finely meshed models that yield a vast number of potential sensor locations. In such cases, the search space becomes large for traditional techniques, such as effective independence. Additionally, in structural health monitoring scenarios that require periodic testing, a learned policy can be incrementally fine-tuned to the structure’s updated FEA model, rather than rerunning a full optimisation with each model revision, thereby reducing computational cost and enabling faster deployment.

Nevertheless, the framework is not without limitations. A notable challenge observed in the current case study is the exponential growth in the number of test environments as the frequency spectrum of interest expands. This growth increases both the computational burden and the training time required to achieve satisfactory policy generalisation. Furthermore, increasing the number of sensors impacts the training efficiency due to the agent's sensor manipulation strategy. Since the agent is constrained to move only one sensor per environment step, the exploration of the configuration space becomes more time-consuming as the total number of sensors increases. This sequential movement model slows convergence in high-dimensional placement scenarios.

Despite the identified limitations, the proposed framework establishes a solid foundation for future research in developing intelligent agent-based test support systems. Future extensions of this work will focus on expanding the framework beyond sensor location to encompass other critical test design decisions. Additionally, to enhance scalability and training efficiency (particularly in high-dimensional design and search spaces), future implementations may incorporate parallel sensor movement strategies, hierarchical policy architectures, and curriculum-aware pruning techniques for the test environment set.

\section{Conclusion}\label{sec: C}
To conclude, traditional sensor placement methodologies in test design are typically static, often centred around a single global modal test configuration. This static approach proves inadequate in test campaigns involving diverse and evolving test configurations, where adaptability is essential to meet specific objectives. In this study, we propose an adaptive, agent-based framework that provides decision support for sensor placement under varying test design parameters, enabling optimal sensor location selection across a range of scenarios. The framework is developed on the underspecified partially observable Markov decision process and solved via a reinforcement-learning agent guided by a dual-curriculum learning strategy. We demonstrate its efficacy on an optimum sensor placement for different segments of the global frequency spectrum, hence facilitating detailed dynamic characterisation of a clamped-cantilever structure.

\section*{Acknowledgement}
 The authors gratefully acknowledge UK Research and Innovation (UKRI) and Siemens Digital Industries Software NV for their support under an EPSRC Industrial Case Award (ref. 2756020). For the purpose of open access, the author(s) has/have applied a Creative Commons Attribution (CC-BY) license  \url{https://creativecommons.org/licenses/by/4.0/} to any Author Accepted Manuscript version arising.

\bibliographystyle{unsrtnat}
\bibliography{references}  %%% Uncomment this line and comment out the ``thebibliography'' section below to use the external .bib file (using bibtex) .





% \end{thebibliography}

\appendix
\addcontentsline{toc}{section}{Appendices}
\renewcommand{\thesubsection}{\Alph{subsection}}

\section{Additional Result}\label{sec: AR}
\begin{figure}[ht]
    \centering
    \includegraphics[width=1.0\linewidth]{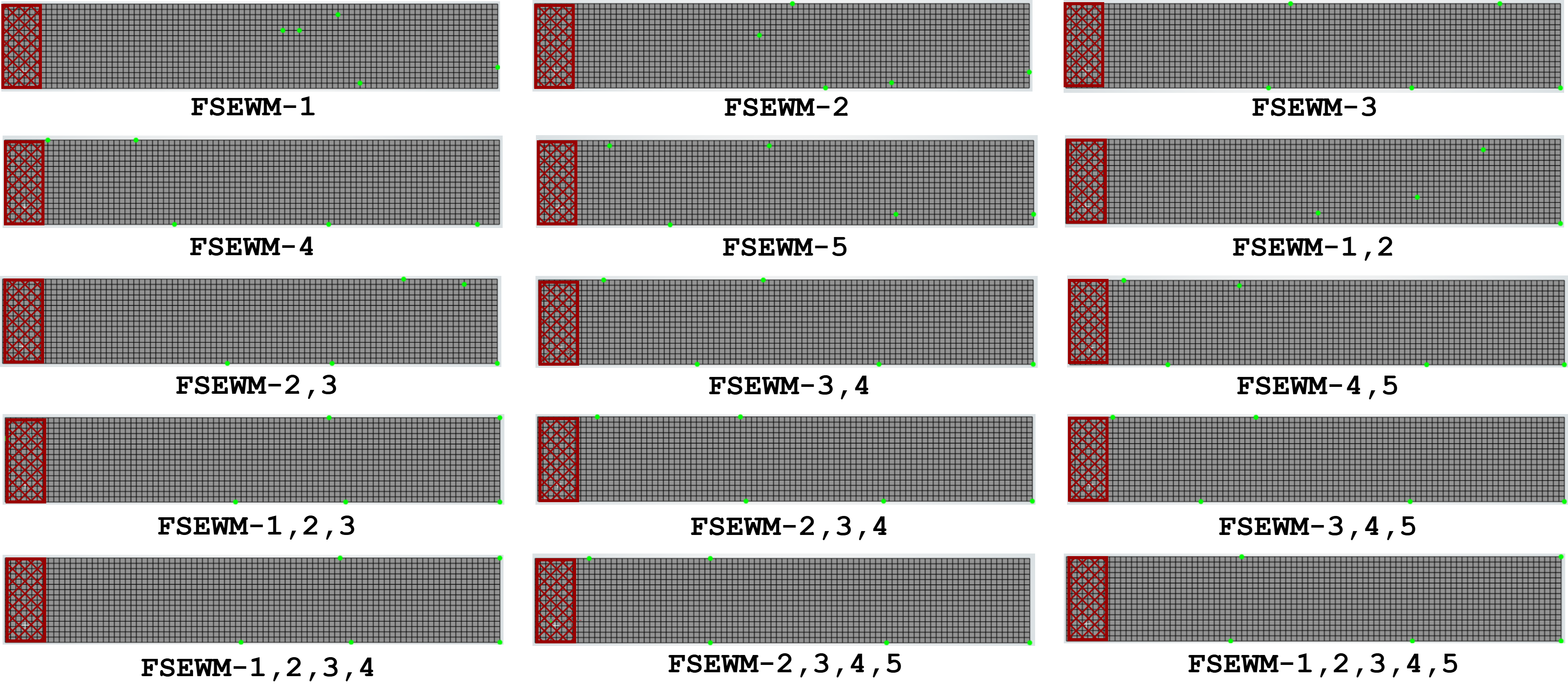}
    \caption{Final sensor configuration prediction from the trained agent across all FSEWM. The green dots indicate the position of all 5 sensors per environment, and the red region is the clamped section of the cantilever plate.}
    \label{fig: Sensor final location}
\end{figure}

\begin{figure}[ht]
    \centering
    \includegraphics[width=1.0\linewidth]{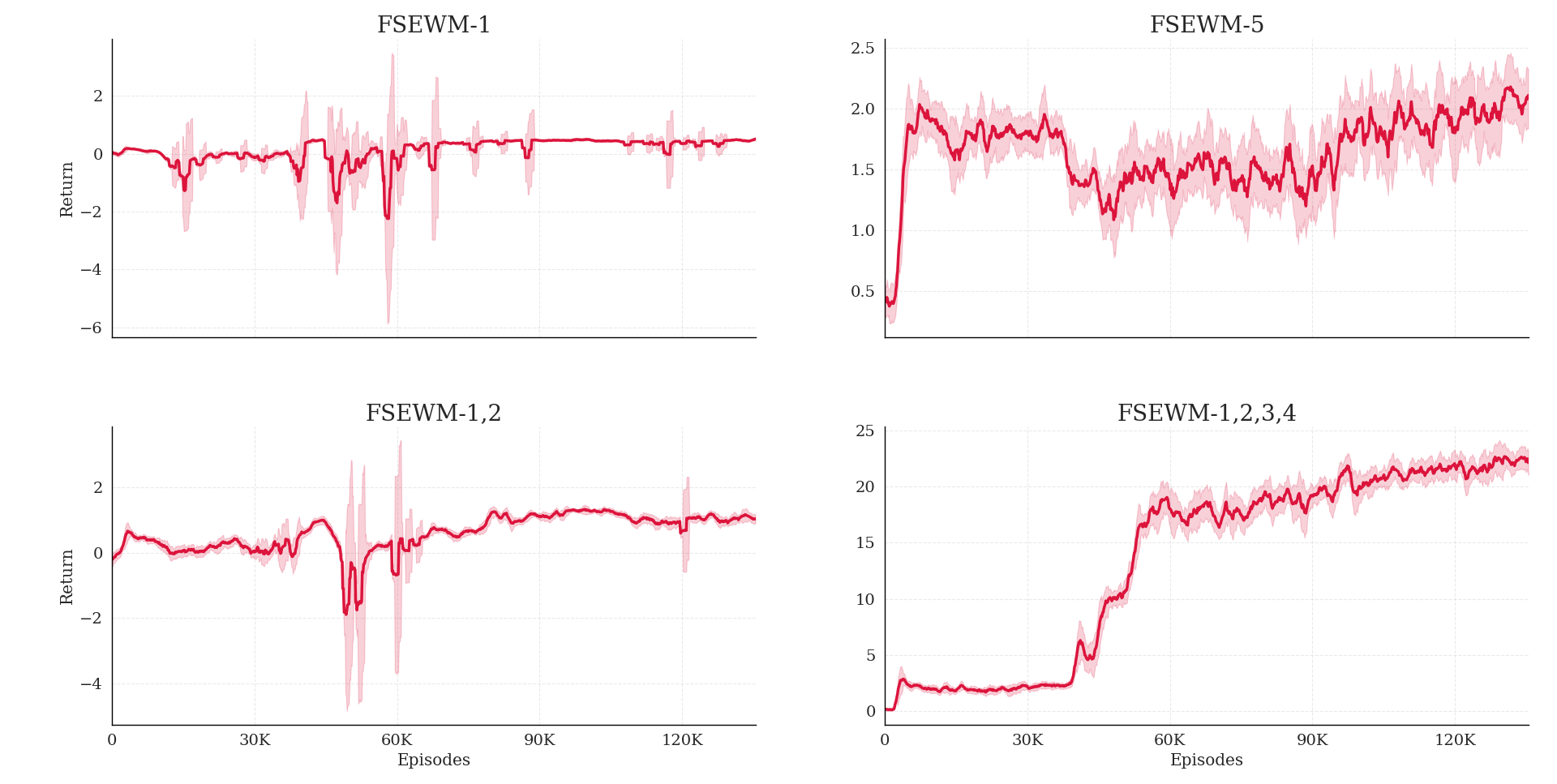}
    \caption{Student agent training performance trajectory on out-of-distribution environments. Curve is smoothed with a moving average of 10 points and standard deviation (shaded region).}
    \label{fig: zero-shot transfer}
\end{figure}

In this section, we present the final sensor configurations selected by the agent across all test environments, as illustrated in Fig. \ref{fig: Sensor final location}. We also visualise how the agent gained performance capability in out-of-distribution environments in Fig. \ref{fig: zero-shot transfer}. This demonstrates that the agent can progressively transfer its learned policy to unseen environments that represent variants of those encountered during training. We hypothesise that this capability emerges from training the agent on a sufficiently diverse subset (specifically, a representative 75\% sample) of the total environment set. During this study, we observed that when the training subset falls below this threshold, the agent's ability to generalise degrades significantly, resulting in diminished performance in unseen environments.

\section{Ablation Study}\label{sec: AS}
\begin{figure}[ht]
    \centering
    \includegraphics[width=1.0\linewidth]{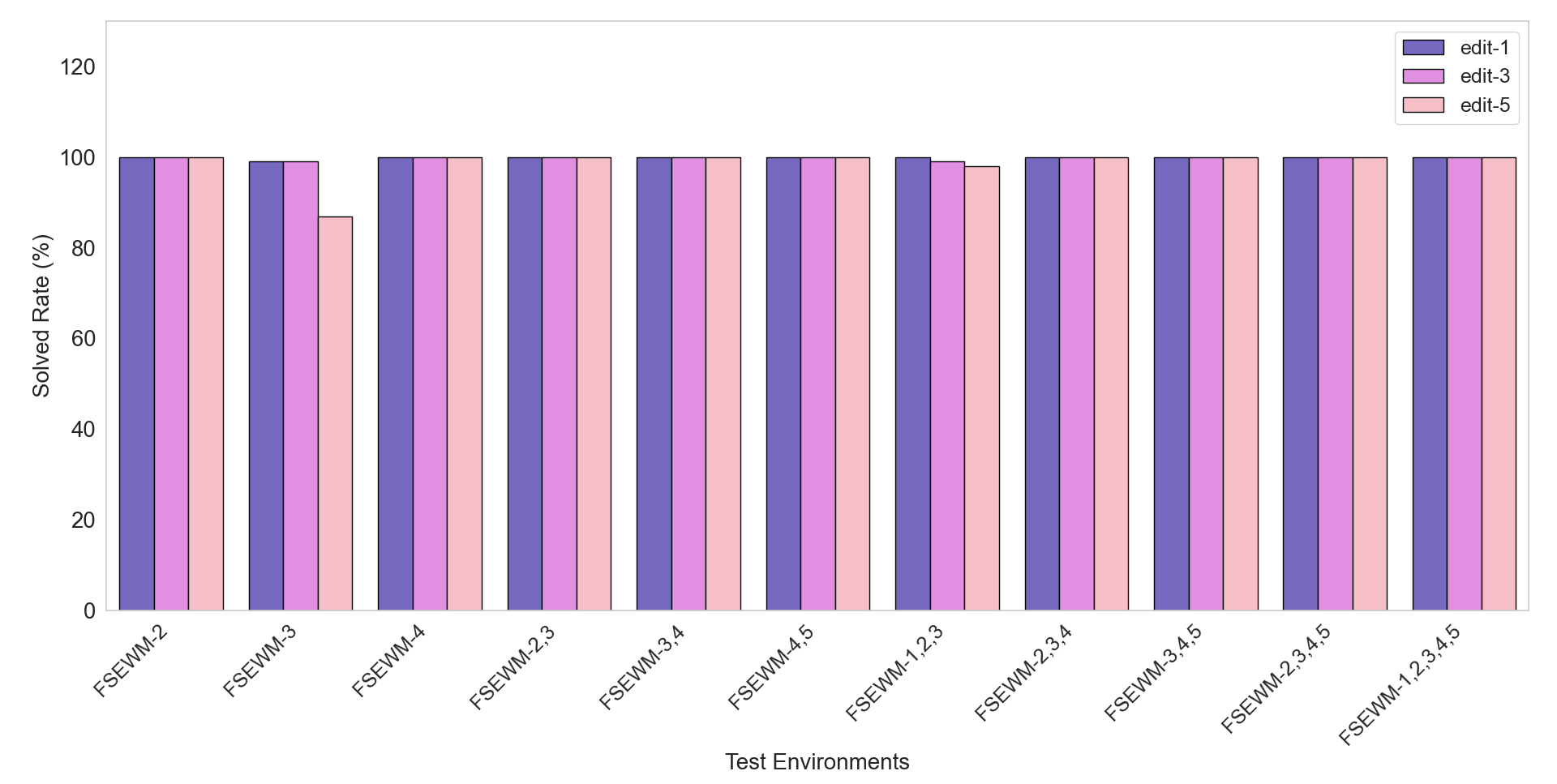}
    \caption{Student agent solved rate performance for ablation study on different numbers of mutated sensors in the trained environments for 100 evaluation episodes.}
    \label{fig: ablation solve rate}
\end{figure}

This section presents an ablation study designed to evaluate the ACCEL mutation technique incorporated in our framework. In particular, we analyse how the number of sensors randomly selected during each mutation cycle impacts the overall performance of the student agent on the trained environments. We conducted the ablation study for sensors within the set $\{1,3,5\}$. The results are presented in Table \ref{tab: ablation study}, which shows that ACCEL with one edit sensor performs better on average across all trained environments. This conclusion is further supported by the solve rate plot in Fig. \ref{fig: ablation solve rate}. Our case study, therefore, adopts a single-sensor edit.

\begin{table}[ht]
\centering
\caption{Student agent mean reward metric performance for ablation study on different numbers of mutated sensors in the trained environments for 100 evaluation episodes.}
\label{tab: ablation study}
\begin{tabular}{lccc}
    \toprule
    \textbf{Test Environments} & \textbf{ACCEL-1 Edit} & \textbf{ACCEL-3 Edit} & \textbf{ACCEL-5 Edit} \\
    \midrule
    FSEWM-2               & $1.74 \pm 0.15$  & $\mathbf{1.82 \pm 0.21}$  & $1.64 \pm 0.16$  \\
    FSEWM-3               & $\mathbf{4.75 \pm 0.33}$  & $4.23 \pm 0.11$  & $4.09 \pm 0.41$  \\
    FSEWM-4               & $2.40 \pm 0.28$  & $\mathbf{2.76 \pm 0.18}$  & $2.34 \pm 0.23$  \\
    FSEWM-2,3             & $\mathbf{12.76 \pm 0.99}$ & $8.50 \pm 0.88$  & $8.45 \pm 0.51$  \\
    FSEWM-3,4             & $5.60 \pm 0.23$  & $11.60 \pm 0.47$  & $\mathbf{11.84 \pm 0.19}$  \\
    FSEWM-4,5             & $5.84 \pm 1.21$  & $\mathbf{6.18 \pm 0.63}$  & $5.75 \pm 0.84$  \\
    FSEWM-1,2,3           & $\mathbf{14.95 \pm 1.35}$ & $7.46 \pm 0.76$  & $7.06 \pm 0.97$  \\
    FSEWM-2,3,4           & $7.91 \pm 0.26$  & $\mathbf{14.74 \pm 0.74}$  & $14.25 \pm 0.25$  \\
    FSEWM-3,4,5           & $\mathbf{10.20 \pm 0.34}$ & $9.91 \pm 0.44$  & $9.57 \pm 0.72$ \\
    FSEWM-2,3,4,5         & $11.30 \pm 0.55$ & $\mathbf{11.50 \pm 0.64}$ & $10.84 \pm 1.04$ \\
    FSEWM-1,2,3,4,5       & $\mathbf{19.89 \pm 1.10}$ & $9.88 \pm 0.51$ & $9.96 \pm 0.56$ \\
    \midrule
    Mean       & $\mathbf{8.21 \pm 0.67}$ & $7.00 \pm 0.46$ & $6.77 \pm 0.58$ \\
    \bottomrule
\end{tabular}
\end{table}

\newpage

 \section{Implementation Details}\label{sec: ID}
Our ACCEL implementation builds on the original codebase developed in \cite{parker2022evolving}, with major adaptations for the problem addressed in this study. It is also important to highlight that each test environment has a unique one-hot identification which is added to the observation space. All training, including the ablation study, was performed on a system equipped with 16.0 GB RAM, an 11th Gen Intel(R) Core(TM) i7-11700 processor, and an Nvidia GeForce GT 730 GPU. The environment was built using the Ansys Pythonic framework, PyAnsys. An open-source implementation of the proposed framework, reproducing our experiments, is available at \url{https://github.com/Collins-Ogbodo/AbTSS.git}.

\newpage
\subsection*{Hyperparameter}\label{subsec: H}
Given the similar discrete structure of both the cantilever environment and the MiniGrid environment, we leveraged several hyperparameters established in \cite{parker2022evolving} to avoid the computational overhead associated with extensive hyperparameter optimisation. A detailed summary of the adopted hyperparameters is provided in Table~\ref{table: hyperparams}.

\begin{table}[ht]
\centering
\caption{Student agent training hyperparameter.}
\begin{tabular}{lc}
    \toprule
    Parameter &  \\
    \midrule
    \multicolumn{2}{l}{\textbf{Environment}} \\[0.5em]
    Episode length              & 200 \\
    Modes               & 1,2,3,4,5 \\
    Mode shape normalisation      & yes \\
    Number of sensors  & 5 \\
    Mesh element size & 5e-3 \\
    \midrule
    \multicolumn{2}{l}{\textbf{PPO}} \\[0.5em]
    $\gamma$               & 0.99 \\
    $\lambda_{\text{GAE}}$ & 0.95  \\
    PPO rollout length     & 256   \\
    PPO epochs             & 5     \\
    PPO minibatches/epoch  & 1     \\
    PPO clip range         & 0.2   \\
    PPO number of workers  & 16    \\
    Adam learning rate     & 1e-4  \\
    Adam $\epsilon$        & 1e-5  \\
    PPO max gradient norm  & 0.5   \\
    PPO value clipping     & yes   \\
    Return normalisation   & yes    \\
    Value loss coefficient & 0.5   \\
    Student entropy coeff. & 0.0   \\
    Generator entropy coeff.& 0.0   \\
    \midrule
    \multicolumn{2}{l}{\textbf{ACCEL}} \\[0.5em]
    Edit rate, $q$         & 1.0   \\
    Replay rate, $p$       & 0.8   \\
    Buffer size, $K$       & 15  \\
    Scoring function       & positive value loss \\
    Edit method            & random \\
    Number of edits        & 1     \\
    Prioritisation         & rank  \\
    Temperature, $\beta$   & 0.3   \\
    Staleness coefficient, $\rho$ & 0.3\\
    \midrule
    \multicolumn{2}{l}{\textbf{LSTM}} \\[0.5em]
    Hidden state  & 256   \\
    \midrule
    \multicolumn{2}{l}{\textbf{PLR}} \\[0.5em]
    Scoring function & positive value loss\\
    Replay rate, p & 0.8\\
    \bottomrule
\end{tabular}\label{table: hyperparams}
\end{table}

\end{document}